\documentclass[10pt,twocolumn,letterpaper]{article}

\usepackage{iccv}
\usepackage{times}
\usepackage{epsfig}
\usepackage{graphicx}
\usepackage{amsmath}
\usepackage{amssymb}
\usepackage{booktabs}
\usepackage{multirow}
\usepackage{makecell}
\usepackage{pifont}
\usepackage{xcolor}

\newlength\savewidth\newcommand\shline{\noalign{\global\savewidth\arrayrulewidth
  \global\arrayrulewidth 1pt}\hline\noalign{\global\arrayrulewidth\savewidth}}

\newcommand{\cmark}{\ding{51}}%
\newcommand{\xmark}{\ding{55}}%

\usepackage[pagebackref=true,breaklinks=true,letterpaper=true,colorlinks,bookmarks=false]{hyperref}

\iccvfinalcopy 

\def\iccvPaperID{2915} 

\ificcvfinal\fi

\begin{document}

\title{PARTNER: Level up the Polar Representation for LiDAR 3D Object Detection}

\author{Ming Nie\textsuperscript{1}\thanks{Equally contributed.}
\quad
Yujing Xue\textsuperscript{2,5}$^*$
\quad
Chunwei Wang\textsuperscript{3}$^*$
\quad
Chaoqiang Ye\textsuperscript{3}
\quad
Hang Xu\textsuperscript{3}
\quad
\\
Xinge Zhu\textsuperscript{4}
\quad
Qingqiu Huang\textsuperscript{4}
\quad
Michael Bi Mi\textsuperscript{5}
\quad
Xinchao Wang\textsuperscript{2}
\quad
Li Zhang\textsuperscript{1}\thanks{Li Zhang (lizhangfd@fudan.edu.cn) is the corresponding author at School of Data Science, Fudan University} \\
\textsuperscript{1}School of Data Science, Fudan University
\quad
\textsuperscript{2}National University of Singapore
\quad \\
\textsuperscript{3}Huawei Noah’s Ark Lab
\quad 
\textsuperscript{4}{Huawei ADS}
\quad
\textsuperscript{5}{Huawei International Pte Ltd}
\quad \\
\vspace{-2mm}
\\
\url{https://github.com/fudan-zvg/PARTNER}
}

\maketitle
\ificcvfinal\thispagestyle{empty}\fi

\begin{abstract}
Recently, polar-based representation has shown promising properties in perceptual tasks.
In addition to Cartesian-based approaches, which separate point clouds unevenly, representing point clouds as polar grids has been recognized as an alternative due to (1) its advantage in robust performance under different resolutions and (2) its superiority in streaming-based approaches.
However, state-of-the-art polar-based detection methods inevitably suffer from the feature distortion problem because of the non-uniform division of polar representation, resulting in a non-negligible performance gap compared to Cartesian-based approaches.
To tackle this issue, we present PARTNER, a novel 3D object detector in the polar coordinate.
PARTNER alleviates the dilemma of feature distortion with global representation re-alignment and facilitates the regression by introducing instance-level geometric information into the detection head.
Extensive experiments show overwhelming advantages in streaming-based detection and different resolutions.
Furthermore, our method outperforms the previous polar-based works with remarkable margins of 3.68\% and 9.15\% on Waymo and ONCE validation set, thus achieving competitive results over the state-of-the-art methods.
\end{abstract}

\section{Introduction}\label{1}
3D object detection serves as an indispensable component in 3D scene understanding for autonomous driving.
With increased affordability and accessibility, LiDAR sensors are now widely adopted for accurate localization.
Unlike regular camera images, LiDAR point cloud has its own unordered, unstructured and sparse nature, thus requiring an effective representation for 3D object detection algorithms in real world.

\begin{figure}[t]
	\begin{center}
		\includegraphics[width=1.0\linewidth]{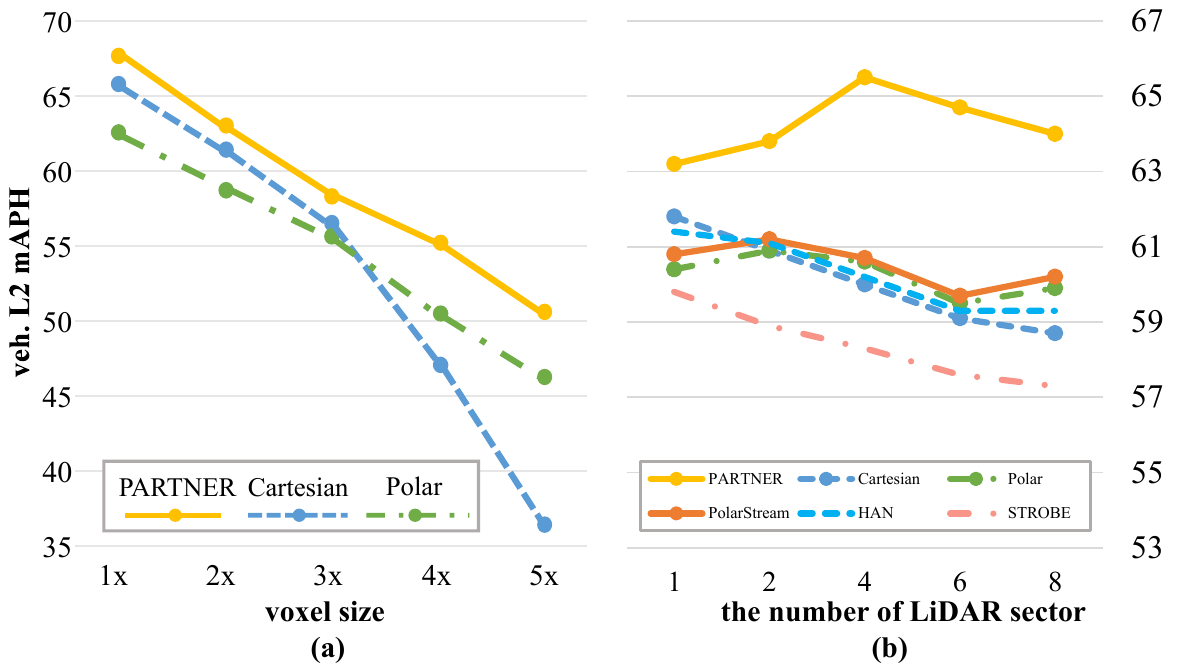}
	\end{center}
	\caption{The superiority of polar representation. Experiments are conducted on Waymo val. set. (a) Performance comparison between different volumetric representations. When enlarging the voxel size of 3D detector CenterPoint~\cite{yin2020center}, replacing cuboid-shaped voxelization with a polar-based one could significantly reduce the performance drop, indicating the accuracy robustness of polar representation.
    (b) Performance in a streaming-based architecture.
    Our PARTNER outperforms the state-of-the-art streaming detector PolarStream~\cite{chen2021polarstream} with remarkable margins.
    }
	\label{fig-resolution}
\end{figure}

Prior arts have explored a variety of representations for point clouds.
Point-based approaches~\cite{shi2019pointrcnn, yang2019std, yang20203dssd} directly process raw points by PointNet and its variants~\cite{qi2017pointnet, qi2017pointnet++}, which also suffer from time-consuming neighbor sampling and grouping operations. 
Range-based methods~\cite{liang2020rangercnn, fan2021rangedet, chai2021point} represent the LiDAR points in a 2.5D manifold structure.
Its compactness enables fast neighbor queries and avoids computation on the empty zone, yet its performance still falls behind other representations.
Alternatively, grid-based methods~\cite{zhou2018voxelnet, lang2019pointpillars, yan2018second,yin2020center,xue2022point2seq} generally transform irregular point clouds into 2D pillars or 3D voxels, and generate 3D boxes in bird's-eye-view (BEV). 
Due to their reasonable computation cost and high performance, these methods have currently dominated the 3D detection benchmark on large-scale datasets~\cite{caesar2020nuscenes, sun2020scalability}. 
Nevertheless, popular Cartesian-based volumetric representation encounters the problem of non-uniform density distribution over distances, where nearby region has much greater density than the distant.
To guarantee affordable memory cost, cells close to sensors have to discard many points during encoding, thus blurring out fine details of nearby objects.
While polar representation shows the ability to relieve this problem, for its volume varying with distance naturally matches the unbalanced distribution of point clouds.
Moreover, its success on LiDAR semantic segmentation task~\cite{zhang2020polarnet, zhu2021cylindrical} motivates us to exploit polar representation in LiDAR 3D object detection.

\begin{figure}[t]
	\begin{center}
		\includegraphics[width=0.9\linewidth]{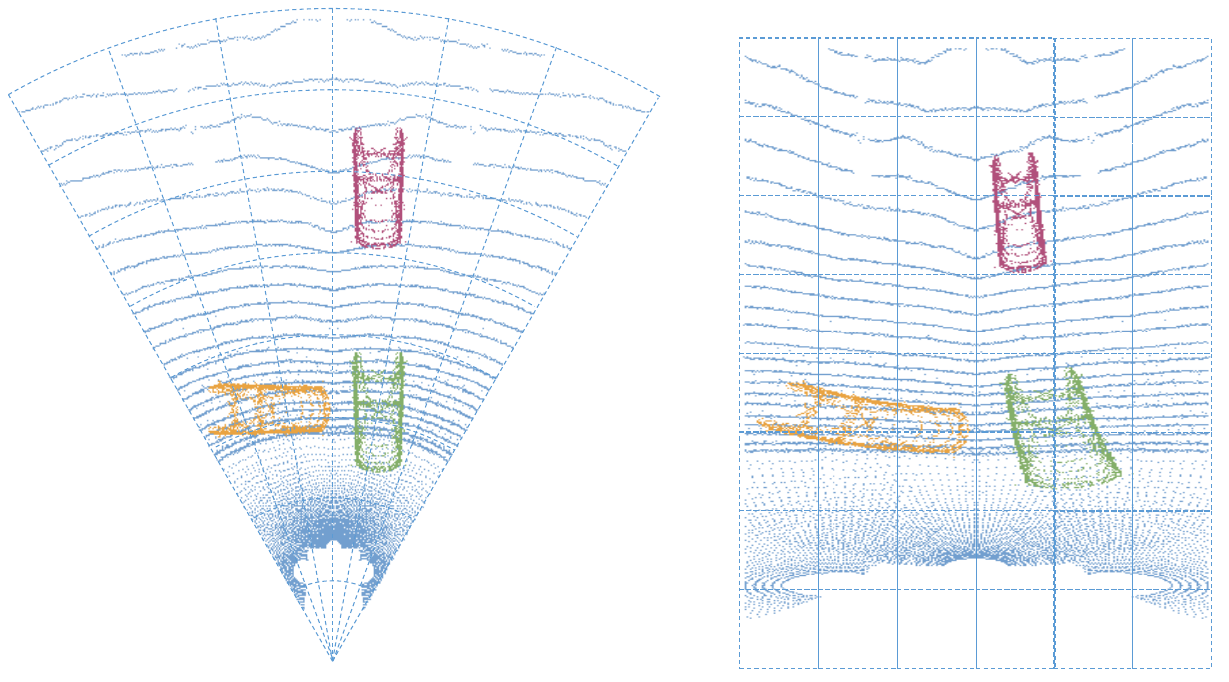}
	\end{center}
\caption{An example of polar feature distortion. Due to the non-uniform division of polar representation, even identical objects in different ranges and headings present diverse distorted appearances, resulting in a global misalignment between objects and regression difficulty for polar-based 3D detectors.}
	\label{fig-distort}
\end{figure}

Polar representation has its potential superiority in several applications.
%
On the one hand, it benefits robust performance under different resolutions due to the consistency between its partition property and the uneven sparsity of point clouds.
As illustrated in Fig.~\ref{fig-resolution}(a), enlarging the voxel size of Cartesian-based 3D detector CenterPoint~\cite{yin2020center} leads to an exponential performance drop, as enlarged voxelization induces more information loss.
In contrast, the accuracy of the polar-based detector decreases linearly due to the non-uniform voxelization process, which aligns with real-world applications in autonomous driving.
%
On the other hand, polar representation naturally resolves the efficiency challenges of existing streaming-based architectures~\cite{han2020streaming, frossard2021strobe, chen2021polarstream}, which processes LiDAR sectors sequentially as soon as scanned data arrives in order to reduce the system latency.
Under this setting, the traditional cuboid-shaped voxelization would inevitably waste both computation and memory on empty voxel regions~\cite{chen2021polarstream}.
In comparison, polar-shaped voxels follow the circular nature of rotating LiDAR emitter and represent wedge-shaped sectors compactly, thus serving as a more suitable solution for streaming-based approaches (shown in Fig.~\ref{fig-resolution}(b)).

Despite its promising characteristics, polar-based 3D detector inevitably suffers from severe feature distortions. 
As illustrated in Fig.~\ref{fig-distort}, identical objects in different ranges and headings present diverse distorted appearances, which reveals two main problems in polar representation.
First of all, due to the salient occupation variations between objects, the translation-invariant convolution is not compatible with non-rectangular structures and is also limited by local receptive fields, which leads to misalignment over global feature maps.
Although PolarStream~\cite{chen2021polarstream} has attempted to address this issue by employing range-stratified convolutions, the cumbersome design fails to re-align the features for feasible object recognition.
%
%
Moreover, traditional CNN-based polar heads undergo difficulties in box prediction.
Owing to the difference in the coordinate system between feature representation and our targeted 3D bounding box, the convolution operation fails to capture the geometric clues in Cartesian coordinates, resulting in regression difficulty.

To address the above issues, we present a novel \textbf{P}ol\textbf{A}r \textbf{R}epresen\textbf{T}atio\textbf{N} det\textbf{E}cto\textbf{R}, dubbed PARTNER, to learn effective representation in polar coordinates.
We introduce two key components, namely global representation re-alignment module and geometry-aware adaptive module.
Global representation re-alignment module employs a novel attention mechanism to re-align the features between objects.
To avoid introducing background noise and efficiently enlarge the receptive field for information interaction, we first condense each column of polar features into a few representative points.
This operation is based on polar-specific observation that features along radial direction only contain limited information due to the occlusion in LiDAR scanning.
After radial feature condensing, angular-wise self-attention is conducted based on those representative points for global feature alignment.
In addition, to facilitate box regression, geometry-aware adaptive module leverages auxiliary supervision to introduce extra geometric and instance information to each pixel. 
Then the predicted information is explicitly introduced into a self-attention mechanism for feature aggregation.
Equipped with the proposed designs, our polar 3D detector achieves superior performance over Cartesian-based methods.

In brief, our contributions are summarized as follows:
\textbf{(i)}
We explore the feature distortion problem in polar representation and propose a novel LiDAR-based 3D object detector, dubbed PARTNER, in the polar coordinate system.
\textbf{(ii)}
We introduce two polar-tailored designs: a global representation re-alignment module to undistort polar-based features and a geometry-aware adaptive module for accurate box prediction.
\textbf{(iii)}
We extensively validate the effectiveness of our designs on the Waymo and ONCE datasets.
Particularly,
our PARTNER achieves competitive results over the state-of-the-art methods and demonstrates overwhelming advantages in streaming and low resolution scenarios.

\begin{figure*}[t]
	\begin{center}
		\includegraphics[width=1.0\linewidth]{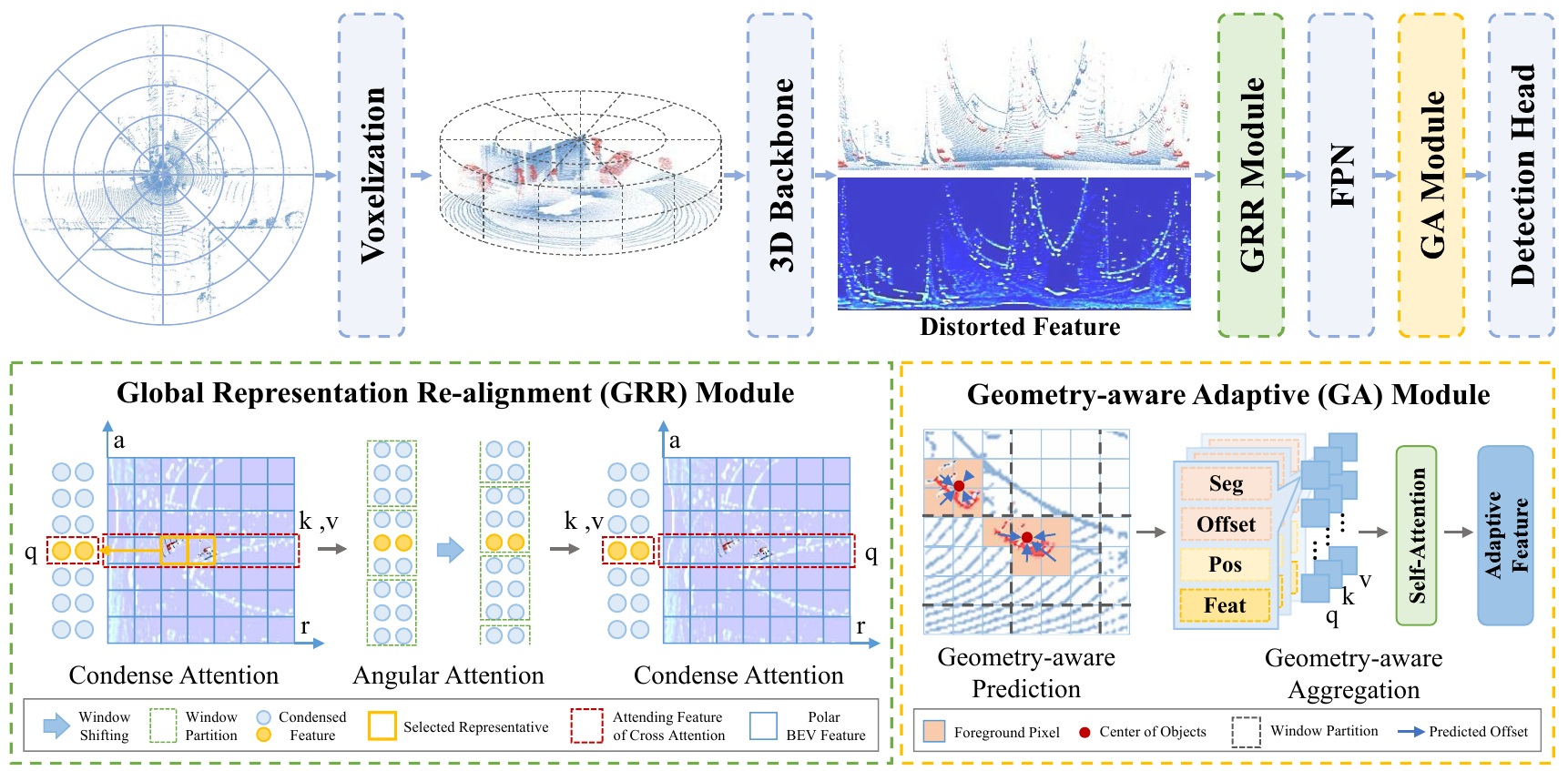}
	\end{center}
	\caption{
The overall architecture of our framework. PARTNER contains four major components: the 3D backbone, the global representation re-alignment module, the 2D FPN, and the geometry-aware adaptive module.
The 3D backbone takes the rasterized point cloud as input and produces the Bird-Eye-View (BEV) feature map for the 3D scene.
The global representation re-alignement module introduces cross-attention between representative features and the corresponding columns as well as self-attention among representative features to re-align the feature representation.
Finally, after the 2D FPN backbone, the geometry-aware adaptive module introduces geometric clues to the feature aggregation process with the help of two auxiliary tasks and detection head produces predicted results.
}
	\label{fig-framework}
\end{figure*}

\section{Related work}
\paragraph{Cartesian-based 3D object detection.}
Current LiDAR-based 3D object detection methods generally transform irregular point clouds to grid representation in Cartesian coordinates and detect objects in bird's-eye-view (BEV).
Pioneer work PointPillars~\cite{lang2019pointpillars} typically groups the points into BEV pillars and employs 2D CNNs to generate 3D proposals. 
In VoxelNet~\cite{zhou2018voxelnet}, points are divided into 3D voxels, and 3D CNNs are leveraged for voxel feature aggregation. 
SECOND~\cite{yan2018second} further proposes a sparse convolution operation for efficient voxel processing. 
CenterPoint~\cite{yin2020center} replaces the general anchor-based detector with a center-based assignment for accurate box localization. 
Until now, grid-based methods dominate the majority of 3D detection benchmarks, especially on large-scale datasets~\cite{sun2020scalability, caesar2020nuscenes}. 


\paragraph{Polar-based 3D object detection.}
Polar or polar-like coordinate system was first explored in LiDAR semantic segmentation tasks. PolarNet~\cite{zhang2020polarnet} and Cylinder3D~\cite{zhu2021cylindrical} point out the varying-density property caused by Cartesian voxels and employ polar-shaped partition to balance the points across grid cells.
However, unlike classification-style tasks, 3D detection additionally requires regression of the object bounding box, thus challenging more on the feature distortion problem in polar representation. %
PolarStream~\cite{chen2021polarstream} proposes a polar-to-Cartesian sampling module for classification and applies range-stratified convolution and normalization to process features over different ranges. 
Considering the limitation of translation-invariant convolution, we resort to the attention mechanism for tackling the challenges of appearance distortion in polar coordinate.

\paragraph{Transformer.}
The success of the transformer has been demonstrated in computer vision tasks. 
Building upon the pioneering work ViT~\cite{dosovitskiy2020image}, many efforts have been devoted to attention mechanism designs.
Axial Attention~\cite{axial} proposes stacked axis-wise transformers for computation efficiency 
while Swin Transformer~\cite{swin} introduces local window attention with window shifting.
FAN~\cite{fan} introduces channel-wise global attention for emerging visual groups of transformers.
Recently, transformer-based architectures are also adopted in LiDAR-based 3D object detection.
VoTr~\cite{votr} and SST~\cite{sst} first apply transformer-based 3D backbones to sparse voxels.
Voxel Set Transformer~\cite{he2022voxset} introduces voxel-based set attention to reduce the self-attention in each voxel.
Compared with previous methods, PARTNER is designed with priors in polar representation to solve the feature distortion issue by learning effective representation. 

\section{Method}
This section presents our \textbf{P}ol\textbf{A}r \textbf{R}epresen\textbf{T}atio\textbf{N} det\textbf{E}cto\textbf{R} (PARTNER), a single stage detector designed for polar BEV representation.
The overall architecture is first introduced in Sec.~\ref{Overall Architecture}.
Then we detail two key components in PARTNER: global representation re-alignment module in Sec.~\ref{rcmodule} and geometry-aware adaptive module in Sec.~\ref{gchead}.

\subsection{Overall architecture} \label{Overall Architecture}
For each 3D scene, PARTNER takes a point cloud consisting of $N_p$ points as input. Each LiDAR point is represented by a vector of point feature $(r_p, a_p, x_p, y_p, z_p, i_p)$, where $(x_p, y_p, z_p)$ is the Cartesian coordinates, $(r_p, a_p)$ is the polar coordinates, and $i_p$ is the reflection intensity.

Fig.~\ref{fig-framework} illustrates the architecture of our proposed \textit{PARTNER}, which consists of four parts:
(1) \textit{The grid-based 3D backbone} leverages grid feature embeddings from points in voxels or pillars and compresses sparse point features into a dense BEV feature map in polar coordinates. 
(2) \textit{Global representation re-alignment module} takes BEV maps as input and produces re-aligned features by applying self-attention to representative features.
(3) \textit{FPN backbone} takes the re-aligned features for feature aggregation and then 
(4) \textit{geometry-aware adaptive module} enhances the feature map with both geometry cues and instance information before producing detection results.

\subsection{Global representation re-alignment module} \label{rcmodule}
In this section, we introduce the Global Representation Re-alignment (GRR) module, a simple yet effective module that applies attention to those essential representative features for global alignment.
The idea behind this design is based on the observation that the spatial resolution of polar grids varies depending on the range.
With such differences, two identical objects in different ranges could occupy a different number of pixels, which results in the feature distortion problem found in existing polar detectors and worsens the performance of the CNN-based feature extractor.

To this end, the global representation re-alignment is designed with two kinds of attention sub-modules:
the condense attention for column feature condensing and the angular attention for feature re-alignment.
Details of two sub-modules will be presented in the following.

\paragraph{Condense attention.} 
As is shown in Fig.~\ref{fig-framework}, the proposed condense attention takes the BEV feature $F \in \mathbb{R}^{R \times A \times C}$ from the 3D backbone as input, where $R$ and $A$ denote the resolution of the radial and azimuth space, and $C$ denotes the number of feature channels. 
In order to select representative features, a 1D local max filtering operation along the radial direction is first applied: 
\begin{equation} \label{3.2.1}
    F^{score}_{sparse} = \mathop{maxfil}\limits_{r = (S, 1)}(F),
\end{equation} 
where $\mathop{maxfil}$ is an operation that seeks out local-maxima in a $S \times 1$ neighborhood, preventing the following selection operation from choosing features representing the same object.
The filtering operation maintains the diversity of the representative features without introducing noise from background pixels.
Then we can get the index of the essential features for each column:
\begin{equation} \label{3.2.2}
    (I_{1}, \cdots, I_{A}) = \mathop{topk}\limits_{dim = 0}(F^{score}_{sparse}),
\end{equation} 
where $I_{i} \in \mathbb{Z}^{N}$ denotes the $N$ indexes of the pixels with the top-k scores in the $i$-th column. 
This top-k operator takes advantage of the data prior: in polar BEV maps, the pixels along the radial direction encode points from a small scanning angle. 
The information in each column is limited due to the occlusion between objects and is distributed in a concentrated manner on the objects or stuff.
Therefore, the selected pixels generally indicate where the 3D object is likely to appear in the 3D scene and could be regarded as representative querying pixels of the column. 
Introducing these representative features avoids the noise and computational overhead introduced by the background area and maintains the diversity of object features in the range aspect. 
Afterward, we compress each column to its representative features.
Let $f^{'}_{i}=F[I_{i}, i, :] \in \mathbb{R}^{N \times C}$  and $f_{i} = F[:, i, :] \in \mathbb{R}^{R \times C}$ be the query and attending features of the $i$-th column, and $p_{i}, p^{'}_{i}$ be the corresponding coordinates of the real pixel centers in both polar and Cartesian systems.
For the $i$-th column, we conduct dot-product attention as follows:
\begin{equation}
    \begin{split}
        & Q_{i} = f^{'}_{i}W_{q}, K_{i} = f_{i}W_{k}, V_{i} = f_{i}W_{v}, \\
        & f^{rep}_{i} = softmax(\frac{Q_{i}K_{i}^T}{\sqrt{d}}) \cdot V_{i}+E(p_{i}, p^{'}_{i}),
    \end{split}
    \label{3.2.3}
\end{equation}
where $W_{q}, W_{k}, W_{v}$ are the linear projection of query, key, and value, $I^{'}_{i}$ is the index of the whole column, and $E(p_{i}, p^{'}_{i})$ is a function of the relative positional encoding:
\begin{equation} \label{3.2.4}
    E(p_{i}, p^{'}_{i}) = ReLU((p_i - p^{'}_{i}) \cdot W_{pos}).
\end{equation}

The cross-attention will be executed for each column to get a condensed feature map $F^{rep} = (f^{rep}_{1}, \cdots, f^{rep}_{A}) \in \mathbb{R}^{N \times A \times C}$ for the following re-alignment operation.

\paragraph{Angular attention.}
With the above condensed representative features, angular attention uses self-attention for global re-alignment.
The design is based on the observation that a distant feature encodes a much larger region than nearby features. 
Thus, it is necessary to introduce an appropriate receptive field in azimuth, which helps distant features access enough nearby features for re-alignment without introducing too much computational overhead. 
With the above prior, we introduce local window attention with window shifting. 
The condensed feature map $F^{rep} \in \mathbb{R}^{N \times A \times C}$ is first divided into non-overlapping windows, where the window size is set to $N \times W_a$. For feature $f_{i}^{w}$ in the $i$-th window, self-attention is applied following Eq.~\ref{3.2.3} and Eq.~\ref{3.2.4}.

After angular attentions, the module applies a reverse condense attention to $f^{r}$ and $F$ to broadcast the re-aligned representative features to each column of the BEV map. 
Let $F^{A} \in \mathbb{R}^{N \times A \times C}$ denote the output feature of the last angular attention.
With $f_{i}= F[:, i, :] \in \mathbb{R}^{R \times C}$ and $f^{a}_{i} = F^{A}[:, i, :] \in \mathbb{R}^{N \times C}$ as the query and attending features respectively, the formulation of the cross-attention is the same as Eq.~\ref{3.2.3} and Eq.~\ref{3.2.4}.

Our model employs two stacked modules for information interaction, where the second angular attention uses a shifted window partition.
Slightly different from the shifting mechanism in Swin Transformer~\cite{swin}, the window is directly rolled without multiplying an attention mask, owing to the existing connection between the most left column and most right column of the polar BEV map.



\begin{figure}[t]
	\begin{center}
		\includegraphics[width=1.0\linewidth]{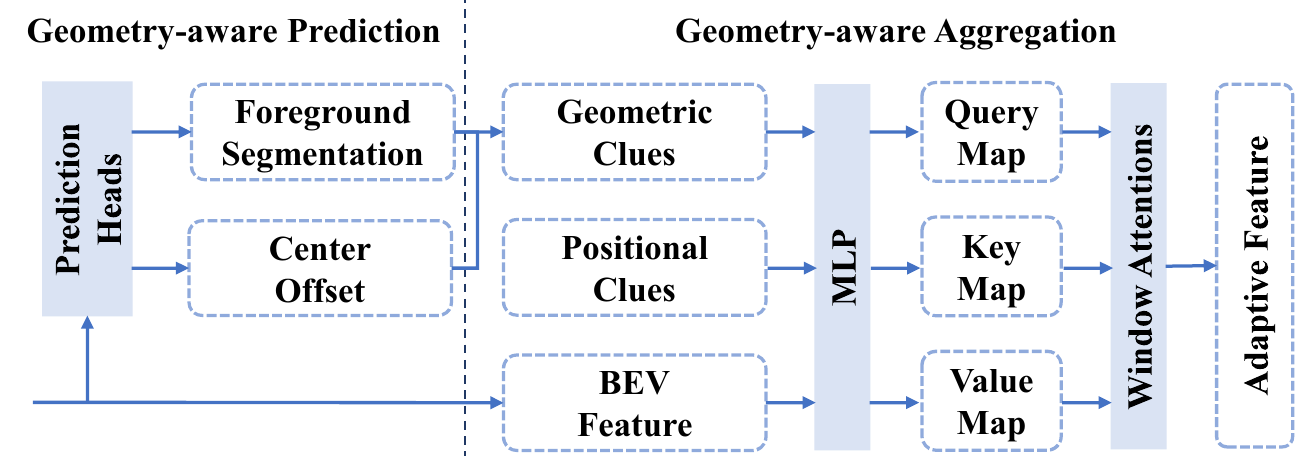}
	\end{center}
	\caption{
The overall architecture of geometry-aware adaptive module.
The geometry-aware prediction module takes 2D BEV features as input, and predicts the foreground segmentation and center regression results.
The geometry-aware aggregation module takes the predicted clues and processes them through an MLP with the positional clues and the BEV features to produce query, key, and value map.
Then two stacked multi-head window attentions are deployed to the maps.
}
	\label{fig-head}
\end{figure}

\subsection{Geometry-aware adaptive module} \label{gchead}
In this section, we first investigate the potential drawbacks of the polar framework on detection heads. 
As is discussed in Sec.~\ref{1}, the spatial resolution of polar grids varies depending on the range, which brings the following problems for the detection head: 1) feature blurring that distant scenes are represented with low resolution, making distant objects close to each other indistinguishable,
and 2) shape variation where traditional CNN-based regression heads tend to fail due to the diverse distorted appearances of objects in different ranges or with different heading angles.
Both of the two issues would lead to regression difficulties in polar detectors.

To address these challenges, we present the Geometry-aware Adaptive (GA) module, a plug-and-play module that is deployed right before the detection head.
The module discriminates features from different instances and explicitly introduces both instance information and geometric clues to the feature aggregation process. 
In the following part, we first introduce the geometry-aware prediction module, followed by the details of the geometry-aware aggregation module.

\paragraph{Geometry-aware prediction.} To obtain fine-grained instance and geometric information, we leverage auxiliary supervisions with two prediction branches: a foreground segmentation branch and a regression branch.
Detailedly, both branches take the BEV features $F^{neck}$ produced by the 2D backbone as input.
The segmentation branch produces a heatmap $H \in \mathbb{R}^{R \times A}$.
Let $\hat{H} \in \mathbb{R}^{R \times A}$ denote the corresponding target map of $H$ and $B_{bev} \in \mathbb{R}^{M \times 5}$ denote the BEV bounding boxes $(x, y, w, h, \theta)$ of the ground-truth objects.
The target map of $\hat{H}$ can be formulated as:
\begin{equation} \label{3.3.1}
    \hat{H}[i] = 
    \begin{cases}
    1 & \text{if } \exists B_{bev}[j], p_{i} \text{ is inside } B_{bev}[j] \\
    0 & \text{otherwise}
    \end{cases},
\end{equation}
where $\hat{H}[i]$ is the $i$-th pixel in $\hat{H}$, and $p_{i}$ is the center coordinate of the $i$-th pixel. 
Then, the branch can be optimized with a focal loss~\cite{Lin_2017_ICCV}:
\begin{equation}
    \mathcal{L}_{fg} = \frac{1}{N_{pos}} \sum^{R \times A}_{i = 1}-\alpha(1-\tilde{H}[i])^{\gamma}log(\tilde{H}[i]),
\end{equation}
where
\begin{equation} \label{3.3.2}
    \tilde{H}[i] = 
    \begin{cases}
    H[i] & \text{if } \hat{H}[i] = 1 \\
    1 - H[i] & \text{otherwise}
    \end{cases}.
\end{equation}

The regression branch predicts the relative distance between the foreground pixels and the centers of their corresponding instances.
Let $D$ and $\hat{D} \in \mathbb{R}^{R \times A \times 4}$ denote the predicted regression map and its corresponding target map, and $C_{bev} \in \mathbb{R}^{M \times 4}$ denote the centers of the ground-truth objects, where each center is represented by both Cartesian coordinate $(x_i, y_i)$ and polar coordinate $(\rho_i, \phi_i)$.
The target map of $\hat{D}$ can be formulated as:
\begin{equation} \label{3.3.3}
    \hat{D}[i] = 
    \begin{cases}
    C_{bev}[k] & \text{if } p_{i} \text{ is inside } B_{bev}[k] \\
    (0, 0, 0, 0) & \text{otherwise}
    \end{cases}.
\end{equation}
For pixels that belong to two or more instances, we randomly choose the corresponding instance it belongs to.
The branch is optimized with a Smooth-$l_1$ loss~\cite{eccv16ssd}:
\begin{equation}
    \mathcal{L}_{dis} = \sum^{R \times A}_{i = 1}\text{Smooth-}l_1(D[i], \hat{D}[i]).
\end{equation}

The output of the module $G = Cat([H, D])$ could be regarded as instance-wise geometric clues, which are supposed to help solve the occupation difference and the feature blurring issue. 
However, we find that the auxiliary supervision itself does not introduce performance gain for polar detectors.
To tackle the issue, we further propose a geometry-aware attention module to fully utilize the predicted information.

\begin{table*}[thb]
    \centering
    \begin{tabular}{l|cc|cc|cc}
    \shline
    \multirow{2}{*}{Method} & \multirow{2}{*}{Backbone} & \multirow{2}{*}{Coordinate} & \multicolumn{2}{c|}{Vehicle LEVEL 1} & \multicolumn{2}{c}{Vehicle LEVEL 2} \\
    & & & 3D mAP(\%) & 3D mAPH(\%) & 3D mAP(\%) & 3D mAPH(\%) \\
    \hline
    \hline
    PointPillars~\cite{lang2019pointpillars} & Pillar & Cartesian & 63.3 & 62.7 & 55.2 & 54.7 \\
    Pillar-OD~\cite{pilarod} & Pillar & Cartesian & 69.8 & - & - & - \\
    MVF~\cite{mvf} & Voxel & Cartesian & 62.93 & - & - & - \\
    PV-RCNN~\cite{pvrcnn} & Voxel & Cartesian & 77.51 & 76.89 & 68.98 & 68.41 \\
    VoTr-TSD~\cite{votr} & Voxel & Cartesian & 74.95 & 74.25 & 65.91 & 65.29 \\
    Pyramid-RCNN~\cite{pyramidrcnn} & Voxel & Cartesian & 76.3 & 75.68 & 67.23 & 66.68 \\
    PolarStream$^{\dag}$~\cite{chen2021polarstream} & Voxel & Polar & 72.37 & 71.81 & 64.56 & 64.04 \\
     \hline
    CenterPoint$^{\dag}$~\cite{yin2020center} & Voxel & Cartesian & 75.58 & 75.01 & 67.00 & 66.52 \\
    CenterPoint$^{\dag}$~\cite{yin2020center} & Voxel & Polar & 72.24 & 71.70 & 63.65 & 63.17 \\
    PARTNER-CP (Ours) & Voxel & Polar & \textbf{76.05} & \textbf{75.52} & \textbf{68.58} & \textbf{68.11} \\
    \hline
    CenterFormer$^{\dag}$~\cite{zhou2022centerformer} & Voxel & Cartesian & 75.83 & 75.32 & 69.52 & 68.98 \\
    CenterFormer$^{\dag}$~\cite{zhou2022centerformer} & Voxel & Polar & 74.79 & 74.30 & 68.56 & 68.04 \\
    PARTNER-CF (Ours) & Voxel & Polar & \textbf{77.76} & \textbf{77.24} & \textbf{70.30} & \textbf{69.84} \\
    \shline
    \end{tabular}
    \caption{Performance comparisons of 3D object detection on the Waymo val. set for vehicle detection.
    We show the mAP/mAPH in the L1/L2 difficulty levels.
    $^{\dag}$: re-implemented using the official code.
	}
    \label{tab:waymo}
\end{table*}

\paragraph{Geometry-aware aggregation.}
This sub-module exploits the self-attention mechanism with the supplement of positional priors of pixels in global coordinates for the following prediction of instance-wise geometric information. 
The proposed module takes the input of the BEV map $F^{neck}$ as well as the output $G$ of the geometry-aware prediction module. 
We first calculate the geometry-aware embedding map $F^{geo}$ from the geometric clues $G$ and the real-world positional clues $P \in \mathbb{R}^{R \times A \times 4}$:
\begin{equation} \label{3.3.4}
    F^{geo} = MLP(Cat([G, P])).
\end{equation} 
Then, we divide $F^{geo}$ and $F^{neck}$ into non-overlapping windows with size $W_g \times W_g$. Let $F^{neck}_{i}, F^{geo}_{i} \in \mathbb{R}^{W_g \times W_g \times C}$ be the feature of $F^{neck}$ and $F^{geo}$ in the $i$-th window, then the query $Q_i$, key $K_i$ and value $V_i$ can be calculated as Eq.~\ref{3.3.5}.

\begin{equation} \label{3.3.5}
\begin{aligned}
Q_{i} = F^{neck}_{i}W_{q} + F_i^{geo}, \\
K_{i} = F^{neck}_{i}W_{k} + F_i^{geo}, \\
V_{i} = F^{neck}_{i}W_{v} + F_i^{geo}.
\end{aligned}
\end{equation}
After that, self-attention can be formulated as Eq.~\ref{3.3.6} and we obtain the aggregated feature of the window $F^{agg}_i$:
\begin{equation} \label{3.3.6}
    F^{agg}_{i} = softmax(\frac{Q_{i}K_{i}^T}{\sqrt{d}}) \cdot V_{i}.
\end{equation}
Similar to the global representation re-alignment module, we stack two geometry-aware aggregation modules with window shifting to enlarge the receptive field.

With the geometry-guided self-attention, the proposed module is capable of leveraging instance and geometric clues into the feature aggregation and learning process, producing features with rich prior for the following heads.

\subsection{Implementation of detection heads}
For simplicity, the design of the task heads follows CenterPoint~\cite{yin2020center}.
To demonstrate the scalability of our method, we also modify the task heads of CenterFormer~\cite{zhou2022centerformer} in the main experiments.
We use smooth-$l_1$ to optimize the predicted box parameters and the focal loss to supervise the classification score.
The hyper-parameters of the task heads just follow \cite{yin2020center} and \cite{zhou2022centerformer}.
Besides, inspired by RangeDet~\cite{fan2021rangedet}, we add an IoU regression head to predict the IoU between the bounding box and the best-matched ground truth annotations, which is also supervised in a smooth L1 loss.
When evaluated, the predicted confidence scores are aligned with regression results by $score' = score * iou^{\alpha}$, where $\alpha$ is set to be $1$ in the experiment.
The overall loss of the proposed head can be formulated as follows:
\begin{equation}
\small
    \mathcal{L} = \mathcal{L}_{cls} + w_{reg}\mathcal{L}_{reg} + w_{fg}\mathcal{L}_{fg} + w_{dis}\mathcal{L}_{dis} + w_{iou}\mathcal{L}_{iou},
\end{equation}
where $w_{reg}$, $w_{fg}$, $w_{dis}$ and $w_{iou}$ are the coefficients that balance the losses and are set to be [2, 1, 0.75, 2] on Waymo models and [0.75, 1, 0.75, 2] on ONCE models.


\section{Experiments}

\begin{table}[t]
\Large
    \renewcommand\arraystretch{1.0}
    \begin{center}
    \resizebox{0.45\textwidth}{!}{
    \setlength{\tabcolsep}{4mm}{
    \begin{tabular}{l|ccccc}
    \shline
    \multirow{2}{*}{Methods} & \multicolumn{5}{c}{Voxel size} \\
    & 1x & 2x & 3x & 4x & 5x \\
    \hline
    \hline
    Cartesian & 65.5 & 61.3 & 56.2 & 47.0 & 35.6 \\
    Polar     & 62.1 & 58.5 & 54.9 & 50.0 & 46.0 \\
    Ours      & \textbf{66.8} & \textbf{62.4} & \textbf{57.0} & \textbf{54.7} & \textbf{50.3} \\
    \shline
    \end{tabular}}
    }
    \end{center}
    \caption{Performance of L2 mAPH under different resolutions (1x to 5x) on the Waymo val. set for vehicle detection.
	}
    \vspace{-10pt}
    \label{tab:resolution}
\end{table}
Experiments are conducted on two large-scale LiDAR object detection benchmarks: Waymo Open Dataset~\cite{sun2020scalability} and ONCE dataset~\cite{2021Once}.
In Sec.~\ref{sec:exp setup}, we introduce the experimental details.
Then we compare our method with other state-of-the-art methods in Sec.~\ref{sec:waymo} and Sec.~\ref{sec:once}.
Finally, to investigate the effect of different components in our method, we conduct ablation studies in Sec.~\ref{sec:ablation}.

\makeatletter 
  \newcommand\figcaption{\def\@captype{figure}\caption} 
  \newcommand\tabcaption{\def\@captype{table}\caption} 
\makeatother
\renewcommand\arraystretch{1.5}

\begin{figure*}
    \begin{minipage}[htbp]{0.48\textwidth}
    \resizebox{1.02\textwidth}{!}{
    \setlength{\tabcolsep}{0.6mm}{
    \begin{tabular}{l|ccccc}
    \shline
    \multirow{2}{*}{Method} & \multicolumn{5}{c}{Number of streaming sectors} \\
    & 1 & 2 & 4 & 6 & 8 \\
    \hline
    \hline
    STROBE~\cite{frossard2021strobe} & 60.5/59.8 & 59.5/58.9 & 58.8/58.3 & 58.3/57.6 & 58.0/57.3 \\
    Han~\cite{han2020streaming} & 61.8/61.4 & 61.7/61.1 & 60.7/60.2 & 60.0/59.3 & 59.9/59.3 \\
    \hline
    Cartesian~\cite{yin2020center} & 62.5/61.8 & 61.6/60.9 & 60.5/60.0 & 59.8/59.1 & 59.4/58.7 \\
    Polar~\cite{yin2020center} & 61.0/60.4 & 61.5/60.9 & 61.0/60.6 & 60.2/59.5 & 60.4/59.9 \\
    \hline
    PolarStream~\cite{chen2021polarstream} & 61.4/60.8 & 61.8/61.2 & 61.2/60.7 & 60.3/59.7 & 60.7/60.2 \\
    Ours & \textbf{63.8/63.2} & \textbf{64.3/63.8} & \textbf{66.0/65.5} & \textbf{65.3/64.7} & \textbf{64.5/64.0} \\
    \shline
    \end{tabular}}}
    \tabcaption{Performance of streaming detection on the Waymo val. set for vehicle detection.
    We show the 6 epochs results in L2.}
    \label{tab:stream_large}
    \end{minipage}
    \hspace{6mm}
    \begin{minipage}[htbp]{0.45\textwidth}
    \includegraphics[scale=0.5]{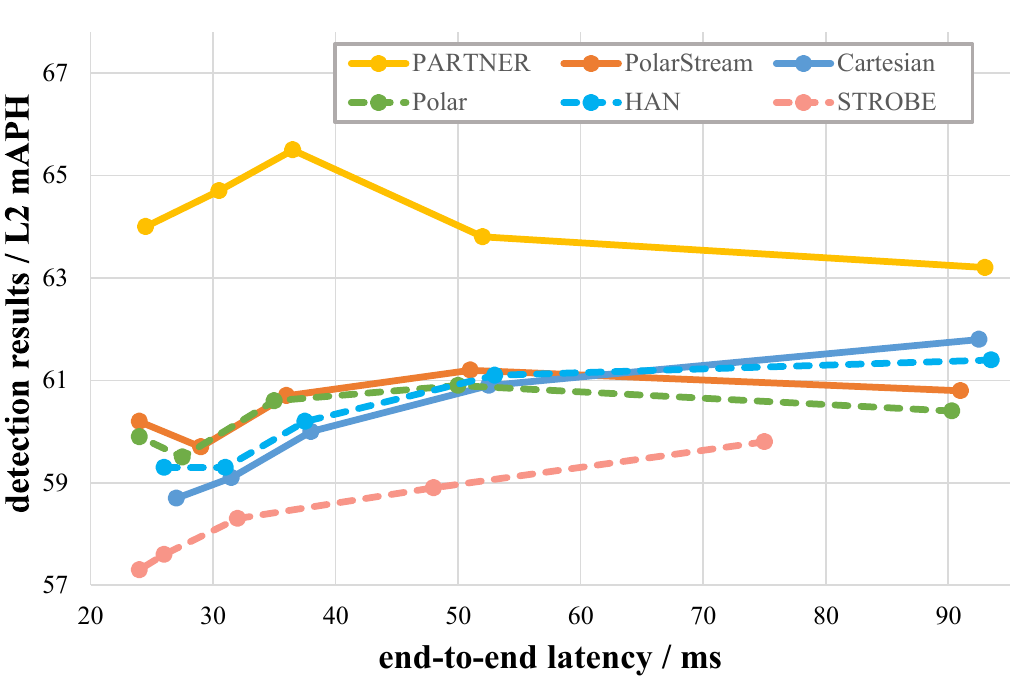}
    \figcaption{Comparison of different streaming methods wrt. detection results vs latency on Waymo val. set.}
    \label{fig:stream_large}
    \end{minipage}
\end{figure*}

\subsection{Experimental setup}
\label{sec:exp setup}
\noindent\textbf{Waymo Open Dataset.}
The Waymo Open Dataset~\cite{sun2020scalability} contains 798 point cloud sequences for training and 202 sequences for validation.
The evaluation metrics on the Waymo Open Dataset are 3D mean Average Precision (mAP) and mAP weighted by heading accuracy (mAPH).
The mAP and mAPH are based on an IoU threshold of 0.7 for vehicles and 0.5 for other categories.
The evaluated detection results are also provided with two difficulty levels: LEVEL 1 for boxes with more than 5 LiDAR points and LEVEL 2 for boxes with at least 1 LiDAR point.
Our model adopts a detection range of [0.3m, 75.18m] for the $r$ axis, [$-\pi$, $\pi$] for the $a$ axis, and [-2m, 4m] for the \textit{z} axis.

\begin{table}[t]
\Large
\renewcommand\arraystretch{1.4}
\resizebox{0.5\textwidth}{!}{
    \centering
    \begin{tabular}{l|c|c|c|c}
    \shline
    Method & Overall & Vehicle & Pedestrian & Cyclist \\
    \hline
    \hline
    PointRCNN~\cite{shi2019pointrcnn} & 28.74 & 52.09 & 4.28 & 29.84 \\
    PointPillars~\cite{lang2019pointpillars} & 44.34 & 68.57 & 17.63 & 46.81 \\
    PV-RCNN~\cite{pvrcnn} & 55.35 & 77.77 & 23.50 & 59.37 \\
    CenterPoint-Cartesian~\cite{yin2020center} & 60.05 & 66.79 & 49.90 & 63.45 \\
    \hline
    CenterPoint-Polar$^{\dag}$~\cite{yin2020center} & 49.55 & 64.62 & 24.67 & 59.35 \\
    PolarStream$^{\dag}$~\cite{chen2021polarstream} & 53.33 & 65.54 & 33.76 & 60.69 \\
    PARTNER (Ours) & \textbf{63.15} & \textbf{68.07} &  \textbf{53.45} & \textbf{65.93} \\
    \shline
    \end{tabular}
    }
    \caption{Performance comparisons of 3D object detection on the ONCE val. set.
    $^{\dag}$: re-implemented using the official code.
	}
	\label{tab:once}
\end{table}

\noindent\textbf{ONCE dataset.}
In the ONCE dataset~\cite{2021Once}, there are one million point cloud frames in total, with 5k, 3k, and 8k frames annotated for training, validation, and testing.
The unlabeled point cloud frames are kept for self-supervised learning or semi-supervised learning.
Our ONCE model is trained on the training set and evaluated on the validation and testing set without using the unlabeled data.
The evaluation metric follows Waymo Open Dataset~\cite{sun2020scalability} and it does not differentiate between L1 and L2.
On the ONCE Dataset, the detection range is set to [0.3m, 75.18m] for the $r$ axis, [$-\pi$, $\pi$] for the $a$ axis, and [-5m, 3m] for the \textit{z} axis.

\noindent\textbf{Implementation details.}
We use the same 3D backbone design, and training schedules as ~\cite{chen2021polarstream, yin2020center}.
The model is trained with the ADAM optimizer and the cosine annealing learning rate scheduler on both two datasets.
On the Waymo Open Dataset, we train our model with batch size 32 and the initial learning rate of 0.003 for 36 epochs on 8 V100 GPUs.
On the ONCE dataset, the batch size and the initial learning rate are not changed, and the training epochs are adapted to 80.
Other training settings follow the respective benchmark.
The voxel size is [0.065, 0.00307, 0.15] for the 1x model.
Due to computational resource limitations, we report the results of streaming-based pipeline, different resolutions, and ablation studies with models trained within 6 or 12 epochs on the Waymo Open Dataset.
More details are included in the appendix.

\subsection{Experiments on the Waymo Open Dataset}
\label{sec:waymo}
\noindent\textbf{Comparison with previous methods.}
For a fair comparison, we re-implement two baseline models from their official implementations: CenterPoint~\cite{yin2020center}, a Cartesian-based method, and PolarStream~\cite{chen2021polarstream}, the state-of-the-art polar-based method.
Our proposed PARTNER, re-implemented PolarStream, and CenterPoint have the same voxel-based 3D backbone, task heads, data augmentations, and training epochs to ensure a fair comparison.
Table~\ref{tab:waymo} shows the detection results on the Waymo validation set.
Adopting CenterPoint~\cite{yin2020center} as baseline, our method attains $76.05\%$ LEVEL 1 mAP and $68.58\%$ LEVEL 2 mAP for vehicle detection, surpassing existing polar-based methods by a significant margin.
Also, our approach outperforms those Cartesian-based 3D detectors~\cite{yin2020center,votr}, manifesting the effectiveness of our method.
Besides, we also report our results with CenterFormer~\cite{zhou2022centerformer} as a stronger baseline.
Experimental results present the state-of-the-art L2 mAP ($70.3\%$), proving the potentiality of polar representation.

\noindent\textbf{Comparison on streaming-based pipeline.}
We perform 3D detection in a streaming manner by separating the point cloud into individual polar sectors and processing each sector similarly to the full-sweep method.
For benchmarking streaming detection, we also re-implement Cartesian representation methods following PolarStream~\cite{chen2021polarstream}.
As is shown in Table~\ref{tab:stream_large}, our method outperforms previous state-of-the-art methods under any number of sectors with a large margin, proving the superiority of our method in steaming-based detection.
Moreover, the trade-offs between latency and accuracy, as illustrated in Figure~\ref{fig:stream_large}, also demonstrate the significant advantages of our method.

To further validate the effectiveness of our PARTNER and its consistency under different grid-based methods, \ie, pillar-based approaches, we deploy our proposed global representation realignment and geometry-aware adaptive head onto PointPillars~\cite{lang2019pointpillars} under streaming scenarios.
As is shown in Table~\ref{tab:stream_pillar}, our method still outperforms PolarStream~\cite{chen2021polarstream} with a pillar-based backbone under any number of sectors with a large margin, indicating both the effectiveness and the generalization of our method.
\begin{table}[ht]
    \Large
    \renewcommand\arraystretch{1.2}
    \begin{center}
    \resizebox{0.48\textwidth}{!}{
    \begin{tabular}{l|cccc}
    \shline
    \multirow{2}{*}{Method} & \multicolumn{4}{c}{Number of streaming sectors} \\
    & 1 & 2 & 4 & 8  \\
    \hline
    \hline
    PolarStream~\cite{chen2021polarstream} & 61.2/60.6 & 60.5/59.9 & 60.3/59.7 & 58.6/58.0 \\
    Ours & \textbf{62.4/61.8} & \textbf{62.8/62.2} & \textbf{62.6/62.0} & \textbf{63.1/62.6} \\
    \shline
    \end{tabular}
    }
    \end{center}
    \caption{Performance of streaming detection based on PointPillars for vehicle detection.
    We show the mAP/mAPH of 6 epochs in the L2 difficulty level for Waymo dataset.
	}
    \label{tab:stream_pillar}
\end{table}

\noindent\textbf{Comparison on different resolutions.}
Another promising intrinsic characteristic of polar-based LiDAR detection is the robust performance under different resolutions.
The experimental results in Table~\ref{tab:resolution} demonstrate that the Cartesian-based method suffers an exponential performance drop (35.6\% in the 5x model), whereas polar-based methods are less affected (50.3\% in the 5x model).
Meanwhile, with the increase of resolution, \ie, the decrease of the grid scale, PolarStream~\cite{chen2021polarstream} shows that it is limited by feature distortion, which is overcome by our method (+4.7\% in 1x model).

\subsection{Experiments on the ONCE dataset}
\label{sec:once}
The ONCE dataset benchmarks different voxel-based detectors using the same backbone network, and we also follow this rule for a fair comparison.
As is shown in Table~\ref{tab:once}, PARTNER attains competitive results in all classes, with $68.07\%$ mAP for vehicle detection, $53.45\%$ mAP for pedestrian detection, and $65.93\%$ for cyclist detection.
The overall mAP of our approach is $63.15\%$, $3.10\%$ higher than the Cartesian-based 3D object detector~\cite{yin2020center}, and $9.82\%$ more elevated than the polar-based 3D object detector~\cite{chen2021polarstream}.
The observations on the ONCE dataset are consistent with those on the Waymo Open Dataset, indicating both the effectiveness and the generalization of our method.

\begin{figure}[t]
	\begin{center}
		\includegraphics[width=1.0\linewidth]{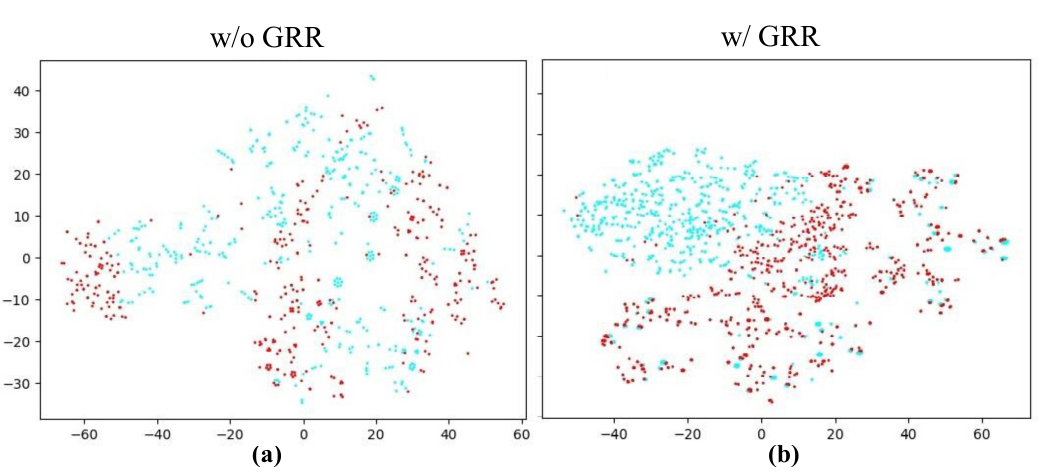}
	\end{center}
	\caption{Visualization of t-SNE features with and without GRR. Blue dots are foreground while red dots are background.}
	\label{fig-tsne}
\end{figure}
\subsection{Ablation studies}
\label{sec:ablation}
\noindent\textbf{Effects of different components in PARTNER.}
\begin{table}[t]
\renewcommand\arraystretch{1.3}
\begin{center}
 \resizebox{1.0\linewidth}{!}{
    \begin{tabular}{l|c|c|p{2.5cm}<{\centering}|p{2.5cm}<{\centering}}
    \shline
    \multirow{2}{*}{Coord.} & \multirow{2}{*}{GRR} & \multirow{2}{*}{GA} & {Veh. LEVEL 1} & {Veh. LEVEL 2} \\
    & & & mAP/mAPH(\%) & mAP/mAPH(\%) \\
    \hline
    \hline
    Cartesian & \xmark & \xmark & 73.51/72.96 & 65.86/65.33 \\
    Polar & \xmark & \xmark & 70.31/69.82 & 62.74/62.23 \\
    Polar & \cmark & \xmark & 72.62/72.07 & 65.02/64.50 \\
    Polar & \cmark & \cmark & \textbf{75.35/74.85} & \textbf{67.84/67.35} \\
    \shline
    \end{tabular}}
    \end{center}
    \caption{Effects of different components in PARTNER.
    We show the 12 epochs results on the Waymo val. set for vehicle detection.
    }
    \label{tab:ablation}
\end{table}
In Table~\ref{tab:ablation}, we investigate the effectiveness of each component in our proposed method.
The global representation re-alignment module can be independently applied on the detector and boost the performance by $2.31\%$ mAP compared to the raw baseline.
Combing the two proposed components, we can obtain a performance gain of $5.04\%$ mAP.

\begin{figure*}[t]
\includegraphics[scale=0.60]{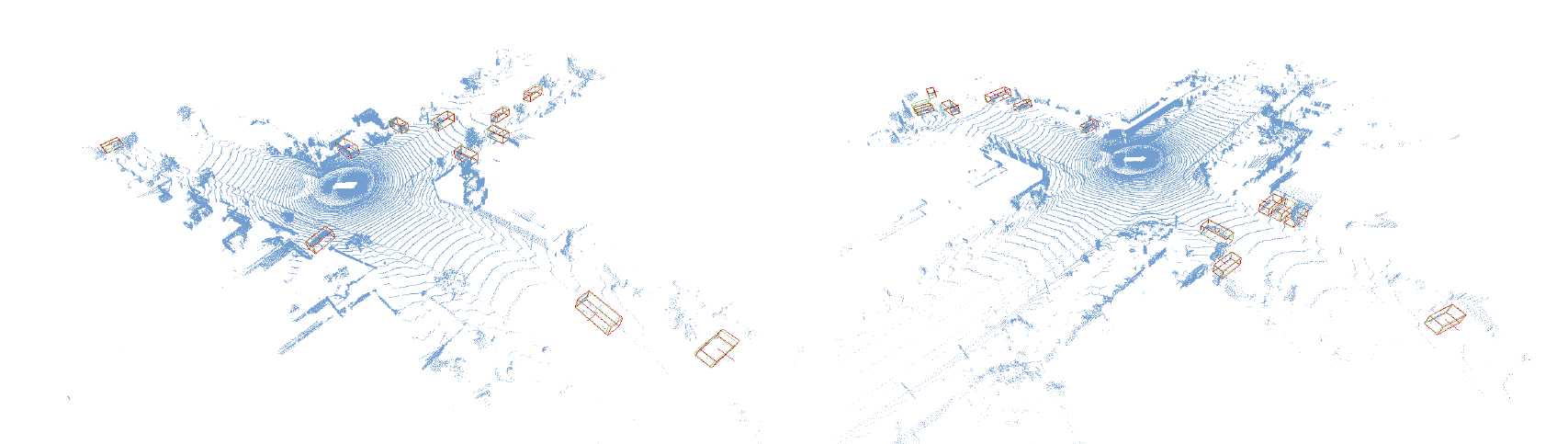}
\caption{Visualization of detection performance on Waymo Open Dataset.
The green box denotes the ground truth box.
The red box denotes the predictions.
Best viewed in color.}
\label{fig:waymo}
\vspace{-7pt}
\end{figure*}

\noindent\textbf{Effects of GRR.}
\begin{table}[t]
\renewcommand\arraystretch{1.3}
\begin{center}
 \resizebox{1.0\linewidth}{!}{
    \begin{tabular}{l|c|p{2.5cm}<{\centering}|p{2.5cm}<{\centering}}
    \shline
    \multirow{2}{*}{Coord.} & \multirow{2}{*}{Re-alignment} & {Veh. LEVEL 1} & {Veh. LEVEL 2} \\
    & & mAP/mAPH(\%) & mAP/mAPH(\%) \\
    \hline
    \hline
    Cartesian & - & 73.51/72.96 & 65.86/65.33 \\
    Cartesian & GRR & 73.04/72.50 & 65.38/64.86 \\
    \hline
    Polar & - & 73.33/72.63 & 64.83/64.23 \\
    Polar (Ours) & GRR & \textbf{75.35/74.85} & \textbf{67.84/67.35} \\
    \shline
    \end{tabular}}
    \end{center}
    \caption{Effects of GRR applied on the Cartesian and polar coordinates respectively.
    We show the 12 epochs results on the Waymo val. set for vehicle detection.
    }
    \vspace{-10pt}
    \label{tab:grr_cart}
\end{table}
In the above part we ablate the effects of global representation re-alignment and demonstrate its significant improvement on polar-based representation.
Another inquiry that might pique our interest is how our GRR will influence the Cartesian-based ones.
To further investigate this question, we apply the GRR module to the detectors based on Cartesian representation.
Interestingly, as is shown in Table~\ref{tab:grr_cart}, Cartesian-based detector with the proposed global representation re-alignment module achieves even lower performance compared to before, which further proves the rationality and pertinence of our method towards polar representation.
Additionally, we visualize the t-SNE of features from GRR.
As is shown in Figure~\ref{fig-tsne}, with the help of GRR, foreground features are better aligned and separated from background features.

\begin{table}[t]
\renewcommand\arraystretch{1.3}
\begin{center}
\resizebox{1.0\linewidth}{!}{
    \begin{tabular}{p{2.5cm}<{\centering}|p{3.2cm}<{\centering}|p{3.2cm}<{\centering}}
    \shline
    \multirow{2}{*}{Components} & {Veh. LEVEL 1} & {Veh. LEVEL 2} \\
    & mAP/mAPH(\%) & mAP/mAPH(\%) \\
    \hline
    \hline
    Baseline & 72.62/72.07 & 65.02/64.50 \\
    + aux. loss & 72.48/71.92 & 64.89/64.36 \\
    + win. attn. & 73.11/72.58 & 65.46/64.95 \\
    + pos. clues & 74.34/73.79 & 66.57/66.06 \\
    + geo. clues & \textbf{75.35/74.85} & \textbf{67.84/67.35} \\
    \shline
    \end{tabular}}
    \end{center}
    \caption{Comparison of different designs in GA module.
    We show the 12 epochs results on the Waymo val. set.
	}
    \vspace{-10pt}
    \label{tab:gc head}
\end{table}
\noindent\textbf{Effects of geometry-aware adaptive module.}
To verify the effectiveness of instance-wise geometric information in geometry-aware adaptive module, we conduct ablation studies in Table~\ref{tab:gc head} to pinpoint where the improvements come from.
We can observe that auxiliary supervision alone does not provide a gain but instead leads to a slight performance drop, which proves that auxiliary loss itself did not lead to improvements.
Meanwhile, the window attention only brings a 0.49\% edge up.
Furthermore, the performance is constantly improved with the gradual introduction of geometric and positional clues by 1.72\% and 1.01\%, indicating the crucial rule of instance-wise geometric information.

\subsection{Qualitative results}
We have compared the quantified results of our PARTNER on two major object detection benchmarks: Waymo Open Dataset~\cite{sun2020scalability} and ONCE dataset~\cite{2021Once}.
In this section, we present our visualization results on Waymo~\cite{sun2020scalability}.
As is shown in Figure~\ref{fig:waymo}, our method could successfully detect objects in the challenging scenes.

\vspace{-7pt}
\section{Conclusions}
In this work, we revisit the problem of feature distortion in the polar representation, which restricts the polar-based 3D detectors severely.
To tackle this issue, we present PARTNER, a novel 3D object detector in the polar coordinate.
PARTNER alleviates the dilemma of feature distortion with global representation re-alignment and facilitates the regression by introducing instance-level geometric information into the detection head.
Results demonstrate that the proposed method outperforms the previous polar-based works with a large margin and achieves competitive results over the state-of-the-art methods.
Furthermore, PARTNER exhibits overwhelming advantages in streaming and different resolutions settings.
We expect our work could inspire future research into polar-based 3D object detection.

\noindent\textbf{Acknowledgments}
This work was supported in part by 
STI2030-Major Projects (Grant No. 2021ZD0200204),
National Natural Science Foundation of China (Grant No. 62106050),
Lingang Laboratory (Grant No. LG-QS-202202-07), Natural Science Foundation of Shanghai (Grant No. 22ZR1407500),
USyd-Fudan BISA Flagship Research Program,
and the National Research Foundation, Singapore under its Medium Sized Center for Advanced Robotics Technology Innovation.
We thank MindSpore for the partial support of this work, which is a new deep learning computing framework.

\bibliographystyle{IEEEtran}

\appendix
\section{Implementation details}
\subsection{3D backbone}
We adopt the same 3D backbone network VoxelNet following CenterPoint~\cite{yin2020center} in our experiments.
For polar voxel representation, we replace the original $x-y$ axis with $r-a$ axis in polar, where $r$ indicates range (distance to the ego in $x$-$y$ axis) and $a$ indicates azimuth.
After voxelization, the normal sparse convolution network is used to extract 3D feature map, which is no different from the Cartesian one.
The 3D backbone downsamples the dimensions of $r$ and $a$ axis with a scale factor of $8$ and the $z$ axis with $16$.
And then the 3D feature map is mapped to the BEV and obtain the BEV feature $F \in \mathbb{R}^{R \times A \times C}$ for the following pipeline.

\subsection{Global representation re-alignment}
The GRR applies attention to the representative features for non-local alignment.
It is composed of two attention sub-modules: Condense attention and Angular attention.
We design Condense attention to select representative features to facilitate long-range re-alignment.
In our experiments, the neighborhood of the 1D local max filtering operation is $3$, and the number of essential features selected from each column ($N$) is set to be $4$.
In Angular attention, the angular window size $W_{a}$ is set to be $8$ and the window shifting stride of the second Angular attention is $4$, which is the half of the angular window size.
All the parameters are guaranteed to be the best configuration.


\begin{table*}[thb]
\Large
\renewcommand\arraystretch{1.4}
\resizebox{\textwidth}{!}{
    \centering
    \begin{tabular}{l|c|cccc|cccc|cccc}
    \shline
    \multirow{2}{*}{Method} & \multirow{2}{*}{mAP(\%)}  & \multicolumn{4}{c|}{Vehicle mAP(\%)} & \multicolumn{4}{c|}{{Pedestrian mAP(\%)}} & \multicolumn{4}{c}{{Cyclist mAP(\%)}}\\
    & & overall & 0-30m & 30-50m & 50m-inf & overall & 0-30m & 30-50m & 50m-inf & overall & 0-30m & 30-50m & 50m-inf \\
    \hline
    \hline
    PointRCNN~\cite{shi2019pointrcnn} & 28.74 & 52.09 & 74.45 & 40.89 & 16.81 & 4.28 & 6.17 & 2.4 & 0.91 & 29.84 & 46.03 & 20.94 & 5.46 \\
    PointPillars~\cite{lang2019pointpillars} & 44.34 & 68.57 & 80.86 & 62.07 & 47.04 & 17.63 & 19.74 & 15.15 & 10.23 & 46.81 & 58.33 & 40.32 & 25.86 \\
    PV-RCNN~\cite{pvrcnn} & 55.35 & 77.77 & 89.39 & 72.55 & 58.64 & 23.50 & 25.61 & 22.84 & 17.27 & 59.37 & 71.66 & 52.58 & 36.17 \\
    CenterPoint-Cartesian~\cite{yin2020center} & 60.05 & 66.79 & 80.10 & 59.55 & 43.39 & 49.90 & 56.24 & 42.61 & 26.27 & 63.45 & 74.28 & 57.94 & 41.48 \\
    \hline
    CenterPoint-Polar$^{\dag}$~\cite{yin2020center} & 49.55 & 64.62 & 77.29 & 59.75 & 38.21 & 24.67 & 30.12 & 20.68 & 11.42 & 59.35 & 71.31 & 55.90 & 31.95 \\
    PolarStream$^{\dag}$~\cite{chen2021polarstream} & 53.33 & 65.54 & 78.03 & 60.13 & 38.76 & 33.76 & 40.94 & 29.57 & 15.32 & 60.69 & 72.53 & 57.10 & 32.34 \\
    PARTNER (Ours) & \textbf{63.15} & \textbf{68.07} & \textbf{79.41} & \textbf{62.77} & \textbf{44.35} & \textbf{53.45} & \textbf{66.56} & \textbf{41.08} & \textbf{26.76} & \textbf{65.93} & \textbf{78.86} & \textbf{62.17} & \textbf{42.23} \\
    \shline
    \end{tabular}
    }
    \caption{Performance comparisons of 3D object detection on the ONCE val. set.
    We maintain the same backbone architecture and training configurations with the baselines on the ONCE benchmark.
    $^{\dag}$: re-implemented using the official code.
    We re-implement CenterPoints in polar coordinates for fair comparison.
	}
	\label{tab:once}
\end{table*}

\section{Detailed experimental results on the ONCE dataset}
In the ONCE dataset~\cite{2021Once}, there are one million point cloud frames in total, with 5k, 3k, and 8k frames annotated for training, validation, and testing.
The unlabeled point cloud frames are kept for self-supervised learning or semi-supervised learning.
Our ONCE model is trained on the training set and evaluated on the validation and testing set without using the unlabeled data.
The evaluation metric follows Waymo Open Dataset~\cite{sun2020scalability} and it does not differentiate between L1 and L2.
On the ONCE Dataset, the detection range is set to [0.3m, 75.18m] for the $r$ axis, [$-\pi$, $\pi$] for the $a$ axis, and [-5m, 3m] for the \textit{z} axis.

The detection results are divided according to the object distances to the sensor: 0-30m, 30-50m, and 50m-inf.
Due to page limitation, in the main paper we only evaluate the performance ignoring the distance.
In this section, we report the detailed performance according to distance in Table~\ref{tab:once}.
Our method attains consistent advantages in different ranges compared to the Cartesian-based 3D object detectors~\cite{shi2019pointrcnn,lang2019pointpillars,yin2020center} as well as polar-based state-of-the-art method~\cite{chen2021polarstream}.

\begin{table}[thbp]
\renewcommand\arraystretch{1.2}
\begin{center}
 \resizebox{0.75\linewidth}{!}{
    \begin{tabular}{l|c|p{2.5cm}<{\centering}}
    \shline
    Methods & Frames & L2 mAPH \\
    \hline
    \hline
    SWFormer & 3 & 74.7 \\
    ConQueR & 1 & 73.3 \\
    PARTNER-CP & 1 & 73.2 \\
    PARTNER-CF & 1 & \textbf{75.0} \\
    \shline
    \end{tabular}}
    \end{center}
    \caption{Results on Waymo test set.
    }
    \label{tab:2xmodelv4}
\end{table}

\section{Experimental results on the Waymo test set}
To make fully comparison, we compare with two recent state-of-the-art works, SWFormer~\cite{sun2022swformer} and ConQueR~\cite{zhu2023conquer} on the Waymo~\cite{sun2020scalability} test set.
Actually, our PARTNER-CF has outperformed SWFormer (single-frame) and ConQueR~\cite{zhu2023conquer} (1x resolution) in L2 on the validation set.
As ConQueR~\cite{zhu2023conquer} only reports its results with 2x resolution backbone (mentioned in ConQueR~\cite{zhu2023conquer}) on the test set, to make comparison fairly, we also utilize the 2x resolution trick and report our results in Tab.~\ref{tab:2xmodelv4}.
Results prove the advantages of our algorithm even compared with state-of-the-art methods.

\section{More ablations}
\subsection{Effects of different partition patterns}
\begin{table}[t]
\renewcommand\arraystretch{1.3}
\begin{center}
 \resizebox{1.0\linewidth}{!}{
    \begin{tabular}{p{2.5cm}<{\centering}|p{3.2cm}<{\centering}|p{3.2cm}<{\centering}}
    \shline
    \multirow{2}{*}{Discretization} & {Veh. LEVEL 1} & {Veh. LEVEL 2} \\
    & mAP/mAPH(\%) & mAP/mAPH(\%) \\
    \hline
    \hline
    SID~\cite{sid} & 74.32/73.69 & 65.81/65.30 \\
    LID~\cite{lid} & 74.61/73.91 & 66.14/65.61 \\
    UD (Ours) & \textbf{75.35/74.85} & \textbf{67.84/67.35}  \\
    \shline
    \end{tabular}}
    \end{center}
    \caption{Comparison of different discretization strategies.
    We show the 12 epochs results on the Waymo val. set.
    }
    \label{tab:discretization}
\end{table}
In the main paper, we first transform the points on Cartesian coordinate system to the polar system by transforming the points $\{(x, y, z)\}$ to points $\{(r, a, z)\}$.
Then the voxelization performs the split on these three dimensions evenly.
Due to polarizing, the evenly partitioned grid will have varying volumes according to range (more farther-away region, larger grid), which coincides with the imbalanced distribution of point clouds.
Nevertheless, partition range evenly may not be the most appropriate setting.
In addition to uniform partition (UD), two widely used range discretization strategies, \ie, spacing-increasing discretization~\cite{sid} (SID) and linear-increasing discretization~\cite{lid} (LID), are also studied.
SID (increasing bin sizes in log space) is defined as
\begin{equation}
    r_{i} = e^{log(r_{min})} + \frac{log(r_{max}/r_{min}) \cdot i}{N_{r}},
\end{equation}
while LID is defined as:
\begin{equation}
    r_{i} = r_{min} + \frac{r_{max} - r_{min}}{N_{r}(N_{r}+1)} \cdot i(i+1),
\end{equation}
where $r_{i}$ is the range value of the i-th range grid index, [$r_{min}$, $r_{max}$] is the full range to be discretized, and $N_{r}$ is the number of grids along the range axis.
As is shown in Table~\ref{tab:discretization}, uniform partition obtains the best performance compared to SID and LID, which indicates the spatial distribution of UD is more consistent with that of point clouds.


\begin{figure*}[t]
\includegraphics[scale=0.60]{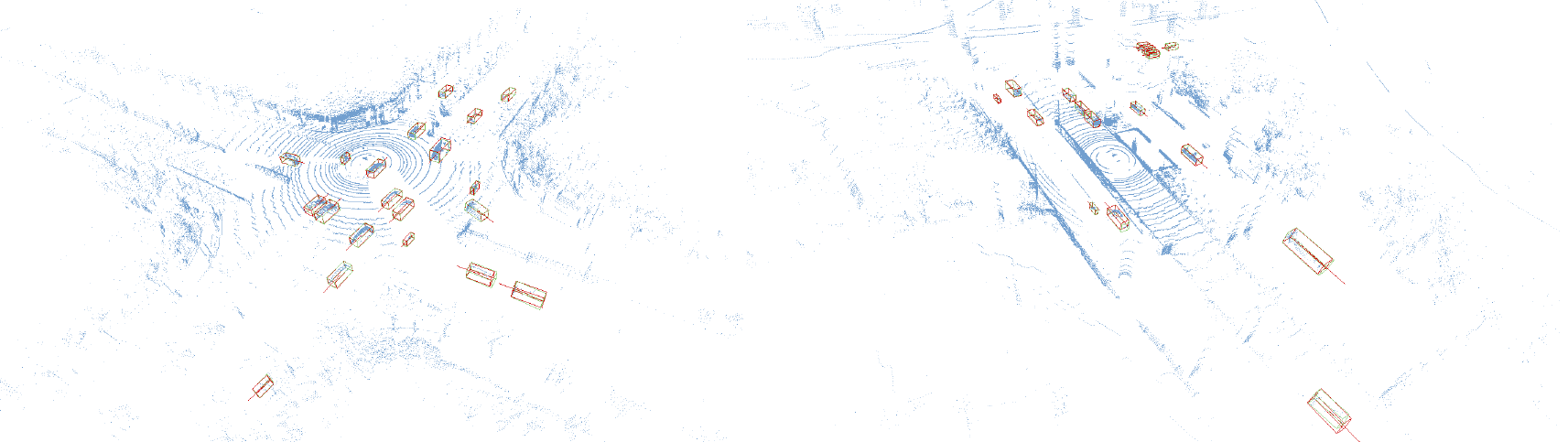}
\caption{Visualization of detection performance on ONCE dataset.
The green box denotes the ground truth box.
The red box denotes the predictions.
Best viewed in color.}
\label{fig:once}
\vspace{-10pt}
\end{figure*}

\subsection{Further study on GRR}
\begin{table}[t]
\renewcommand\arraystretch{1.3}
\begin{center}
 \resizebox{1.0\linewidth}{!}{
    \begin{tabular}{p{3.2cm}<{\centering}|p{3.2cm}<{\centering}|p{3.2cm}<{\centering}}
    \shline
    \multirow{2}{*}{Re-alignment} & {Veh. LEVEL 1} & {Veh. LEVEL 2} \\
    & mAP/mAPH(\%) & mAP/mAPH(\%) \\
    \hline
    \hline
    Baseline & 73.33/72.63 & 64.83/64.23  \\
    Axial~\cite{axial} & 73.41/72.72 & 64.92/64.33 \\
    Channel-wise~\cite{fan} & 73.71/73.00 & 65.22/64.61 \\
    Swin~\cite{swin} & 73.62/72.93 & 65.13/64.54 \\
    GRR & \textbf{75.35/74.85} & \textbf{67.84/67.35} \\
    \shline
    \end{tabular}}
    \end{center}
    \caption{Comparison of different attention mechanisms in the re-alignment.
    We show the 12 epochs results on the Waymo val. set.
    }
    \label{tab:condense}
\end{table}
\begin{table}[t]
\renewcommand\arraystretch{1.3}
\begin{center}
 \resizebox{1.0\linewidth}{!}{
    \begin{tabular}{l|c|p{2.5cm}<{\centering}|p{2.5cm}<{\centering}}
    \shline
    \multirow{2}{*}{Coords.} & \multirow{2}{*}{Reg. Target} & {Veh. LEVEL 1} & {Veh. LEVEL 2} \\
    & & mAP/mAPH(\%) & mAP/mAPH(\%) \\
    \hline
    \hline
    Cartesian & Cartesian & 73.51/72.96 & 65.86/65.33 \\
    \hline
    Polar & Polar & 75.02/74.55 & 67.52/67.04 \\
    Polar (Ours) & Cartesian & \textbf{75.35/74.85} & \textbf{67.84/67.35} \\
    \shline
    \end{tabular}}
    \end{center}
    \caption{Ablation study on the coordinate of box regression target.
    We show the 12 epochs results on the Waymo val. set for vehicle detection.
    }
    \label{tab:regress}
\end{table}
In Table~\ref{tab:condense}, we evaluate different attention mechanisms used for feature re-alignment.
An attempt to enlarge the receptive field efficiently is to integrate polar feature maps globally along the range and angular axis using axial attention~\cite{axial}.
However, experimental results demonstrate this method is not feasible (only $0.08\%$ gain).
Another attempt is to introduce channel-wise attention~\cite{fan} during polar feature re-alignment, by which distorted representations can also be integrated efficiently in a global manner.
Unfortunately, this practice attains limited improvement with $0.38\%$ gain.
One reasonable explanation for the failure of both approaches is that conducting alignment blindly at the global level introduces more background noise and produces feature representations with low quality.
Meanwhile, we try to use swin attention~\cite{swin} to carry out feature alignment within the local region, trivial performance proved by the ablations ($0.29\%$ gain), limited by background noise and restricted receptive field.

\subsection{Study on different regression targets}
In this section, we ablate different regression coordinates.
In our main paper, we regress targets in the Cartesian coordinates, \ie, [$x$, $y$, $z$, $w$, $l$, $h$].
As we conduct feature representation in polar space, a reasonable attempt is to regress targets also in polar coordinates.
We perform this experiment in Table~\ref{tab:regress}.
The experimental results show that regression in polar coordinates is less proper.
One reasonable explanation for this phenomenon is that polar-based representation is robust enough to regress targets in the real world, while regression in polar will bring unnecessary difficulties in optimization and convergence.
For example, the azumith's restricted range of 0 to 2$\pi$ is a numerical obstacle for regression but significantly impacts box prediction results.
We observe a performance degradation, which coincides with our analysis.


\section{Qualitative results}
We have compared the quantified results of our PARTNER on two major object detection benchmarks: Waymo Open Dataset~\cite{sun2020scalability} and ONCE dataset~\cite{2021Once}.
In this section, we present our visualization results on ONCE dataset~\cite{2021Once}.
As is shown in Figure~\ref{fig:once}, our method could successfully detect objects in the challenging scenes and make considerably accurate predictions.

\end{document}